\def\printing{myijcai}
\def\myijcai{myijcai}
\def\blind{blind}
\def\article{article}
\def\frontpage{frontpage}
\ifx\printing\myijcai\documentclass{article} \usepackage{ijcai03}
   \else\ifx\printing\blind\documentclass{article} \usepackage{ijcai03}\else
   \documentclass{article}\fi\fi
\usepackage{times}
\usepackage{epic}
\usepackage{ecltree}
\usepackage{graphicx}
\usepackage{latexsym}

\begin{document}

\newlength{\halftextwidth}
\setlength{\halftextwidth}{0.47\textwidth}
\def\halffigsize{2.2in}
\def\thirdfigsize{1.5in}
\def\negvspace{0in}
\def\posvspace{0em}

\input epsf




\newcommand{\deriv}[2]{\Delta #1_#2}

\newcommand{\tighter}[2]{\mbox{$#1 \preceq #2$}}
\newcommand{\stighter}[2]{\mbox{$#1 \prec #2$}}
\newcommand{\ac}[1]{\mbox{$ac(#1)$}}
\newcommand{\acmin}[1]{\mbox{$ac_{\min}(#1)$}}
\newcommand{\pc}[1]{\mbox{$pc(#1)$}}
\newcommand{\spc}[1]{\mbox{$spc(#1)$}}
\newcommand{\incomparable}[2]{\mbox{$#1 \sim #2$}}
\newcommand{\twodarray}[4]{\mbox{\scriptsize $\left( \hspace{-0.5em} \begin{array}{cc} #1 & #2 \\ #3 & #4 \end{array} \hspace{-0.5em} \right)$}}

\newcommand{\threedarray}[9]{\mbox{\scriptsize $\left( \hspace{-0.5em} \begin{array}{ccc} #1 & #2 & #3 \\ #4 & #5 & #6 \\ #7 & #8 & #9  \end{array} \hspace{-0.5em} \right)$}}

\renewcommand{\theenumii}{\alph{enumii}}
\renewcommand{\theenumiii}{\roman{enumiii}}
\newcommand{\figref}[1]{Figure \ref{#1}}
\newcommand{\tref}[1]{Table \ref{#1}}
\newcommand{\size}{\mbox{$N$}} 
\newcommand{\prob}[1]{\mbox{{P}r\{\scriptsize#1\}}}
\newcommand{\probB}[1]{\mbox{{P}r\{#1\}}}
\newcommand{\secref}[1]{Section \ref{#1}}
\newcommand{\myfrac}[2]{(#1)/#2}
\newcommand{\CSP}{\mbox{\sc Csp}}
\newcommand{\Csp}{\mbox{\sc Csp}}
\newcommand{\SAT}{\mbox{\sc Sat}}

\def\op{\mbox{$\kappa$}}
\def\opcrit{\mbox{$\kappa_{c}$}}
\def\opn{\mbox{$\gamma$}}
\newcommand{\nmpp}{\mbox{$\langle n,m,p_{1},p_{2}\rangle$}}
\newtheorem{mytheorem}{Theorem}
\newtheorem{mytheorem1}{Theorem}
\newcommand{\myproof}{\noindent {\bf Proof:\ \ }}
\newcommand{\myqed}{\mbox{QED.}}

\title{Scenario-based Stochastic Constraint Programming}
\ifx\printing\blind
\author{
Content areas: constraint programming, constraint satisfaction\\
Tracking number: E0036}
\date{}
\else \fi

\ifx\printing\myijcai
\author{Suresh Manandhar \and Armagan Tarim \\
Department of Computer Science \\
University of York, England \\
email: \{suresh,at\}@cs.york.ac.uk \And
Toby Walsh\\
Cork Constraint Computation Centre \\
University College Cork, Ireland. \\
email: tw@4c.ucc.ie.
}
\else \fi

\ifx\printing\article
\author{Suresh Manandhar \and Armagan Tarim \\
Department of Computer Science \\
University of York, England \\
email: \{suresh,at\}@cs.york.ac.uk \And
Toby Walsh\\
Cork Constraint Computation Centre \\
University College Cork, Ireland. \\
email: tw@4c.ucc.ie. }
\date{1st Jan 2003}
\else \fi

\ifx\printing\frontpage
\author{Suresh Manandhar \and Armagan Tarim \\
Department of Computer Science \\
University of York, England \\
email: \{suresh,at\}@cs.york.ac.uk \And
Toby Walsh\\
Cork Constraint Computation Centre \\
University College Cork, Ireland. \\
email: tw@4c.ucc.ie. }

\else \fi

\maketitle
\begin{abstract}
To model combinatorial decision problems involving uncertainty
and probability, we extend the stochastic constraint programming framework proposed in \cite{wecai2002} along a number of important dimensions (e.g. to multiple
chance constraints and to a range of new objectives).
We also provide a new (but equivalent) semantics
based on scenarios. Using this semantics, we can compile stochastic
constraint programs down into conventional (non-stochastic)
constraint programs. This allows us to exploit
the full power of existing constraint solvers.
We have implemented
this framework for decision making under uncertainty
in stochastic OPL, a language which is based on
the OPL constraint modelling language
\cite{hent1}. To illustrate the potential of this
framework, we model a wide range of problems
in areas as diverse as finance, agriculture and
production.
\end{abstract}

\ifx\printing\frontpage
\begin{center}
\begin{tabular}{ll}
Content areas: & constraint programming, constraint satisfaction, reasoning under uncertainty\\
Tracking number: & 142 \\
\end{tabular}
\end{center}

\begin{quote}
This paper has not already been accepted by and is not currently under review for
a journal or another conference, nor will it be submitted for such during IJCAI's
review period.
\end{quote}

\else \fi

\ifx\printing\frontpage
\eject \end{document} \else \fi

\section{Introduction}

Many 
decision problems contain uncertainty. Data about events in the
past may not be known exactly due to errors in measuring or
difficulties in sampling, whilst data about events in the future
may simply not be known with certainty. For example, when
scheduling power stations, we need to cope with uncertainty in
future energy demands. As a second example, nurse rostering in an
accident and emergency department requires us to anticipate
variability in workload. As a final example, when constructing a
balanced bond portfolio, we must deal with uncertainty in the
future price of bonds. To deal with such situations,
\cite{wecai2002} has proposed an extension of constraint
programming, called {\em stochastic constraint programming}, in
which we distinguish between decision variables, which we are free
to set, and stochastic (or observed) variables, which follow some
probability distribution. This framework combines together some of
the best features of traditional constraint satisfaction,
stochastic integer programming, and stochastic satisfiability.

In this paper, we extend the expressivity of this framework
considerably by adding multiple chance constraints, as well as a
range of objective functions like maximizing the downside. We show
how such stochastic constraint programs can be compiled down into
conventional (non-stochastic) constraint programs using a
scenario-based interpretation. This compilation allows us to use
existing constraint solvers without any modification, as well as
call upon the power of hybrid solvers which combine constraint
solving and integer programming techniques. We also propose a
number of techniques to reduce the number of scenarios and to
generate robust solutions. We have implemented this framework for
decision making under uncertainty in a language called stochastic
OPL. This is an extension of the OPL constraint modelling language
\cite{hent1}. Finally, we describe a wide range of problems that
we have modelled in stochastic OPL that illustrate some of its
potential.

\section{Stochastic constraint programs}

In a one stage stochastic constraint satisfaction problem
(stochastic CSP), the decision variables are set before the
stochastic variables. The stochastic variables, independent of the decision variables,
take values with probabilities given by a 
probability distribution.
This models situations
where we act now and observe later.
For example, we have to decide now which nurses to
have on duty and will only later discover the actual workload.
We can easily invert the instantiation order if
the application demands, with the stochastic variables
set before the decision variables. 
Constraints are defined (as in traditional constraint
satisfaction) by relations of allowed tuples of values.
Constraints can, however, be implemented with specialized
and efficient algorithms for consistency checking.

We allow for both hard constraints which are always
satisfied and ``chance constraints'' which may only be satisfied
in some of the possible worlds. Each chance constraint has a
threshold, $\theta$ and the constraint must be satisfied in at least a fraction $\theta$
of the worlds. A one stage stochastic CSP is satisfiable
iff there exists values for the decision variables so that, given
random values for the stochastic variables, the hard constraints
are always satisfied and the chance constraints are satisfied in
at least the given fraction of worlds. Note that \cite{wecai2002}
only allowed for one (global) chance constraint so the definition
here of stochastic constraint programming is strictly more
general.

In a two stage
stochastic CSP,
there are two sets of decision variables, $V_{d1}$ and
$V_{d2}$, and two sets of stochastic variables,
$V_{s1}$ and $V_{s2}$. The aim is to find
values for the variables in
$V_{d1}$, so that given random
values for $V_{s1}$, we can find
values for $V_{d2}$, so that
given random  values for $V_{s2}$,
the hard constraints are always satisfied and
the chance constraints are
again satisfied in at least
the given fraction of worlds.
Note that the values chosen for the second
set of decision variables $V_{d2}$ are conditioned
on both the values chosen for
the first set of decision variables $V_{d1}$
and on the random values given to the first
set of stochastic variables $V_{s1}$.
This can model situations in which
items are produced and can be consumed
or put in stock for later consumption.
Future production then depends both
on previous production (earlier decision
variables) and on previous demand (earlier
stochastic variables).

An $m$ stage stochastic CSP is defined in an analogous way to one
and two stage stochastic CSPs. Note that \cite{wecai2002} insisted
that the stochastic variables take values independently of each
other.
This prevents us representing a number of common
situations. For example, if the market goes down in
the first quarter, it is probably more likely to go down
in the second quarter. A second stage stochastic variable representing
the market index is therefore dependent on the
first stage stochastic variable representing
the market index. There is, however, nothing in
the semantics given for stochastic constraint programs
nor in the solution methods proposed in
\cite{wecai2002} that used this assumption. We
therefore allow later stage stochastic variables
to take values which are conditioned by the earlier
stage stochastic variables.

A stochastic constraint optimization problem (stochastic COP)
is a stochastic CSP plus
a cost function defined over the decision and stochastic variables.
In \cite{wecai2002}, the only goal considered
was to find a solution that satisfies
the stochastic CSP which minimizes or maximizes
the expected value of the objective function.
We now extend this to a much wider range of goals.
For example, we might wish to limit the downside (i.e. maximize
the least value of the cost function),
or to minimize the spread
(i.e. minimize the difference between the least and the
largest value of the cost function).

\section{Scenario-based semantics}

In \cite{wecai2002}, a semantics for
stochastic constraint programs is given based on policies.
A policy is a tree of decisions. Each path in a policy
represents a different possible scenario (set of
values for the stochastic variables), and the
values assigned to decision variables in this
scenario. To find satisfying policies, \cite{wecai2002} presents
backtracking and forward checking algorithms
which explores
the implicit AND/OR graph. Stochastic variables
give AND nodes as we must find a policy
that satisfies all their values, whilst
decision variables give OR nodes as we only
need find one satisfying value.

An alternative semantics, which suggests an alternative solution method, comes from a scenario-based view \cite{birge1}. 
A scenario is any possible set of
values for the stochastic variables.
Thus, a scenario is associated with each
path in the policy.
Within each scenario, we have a conventional
(non-stochastic) constraint program to solve.
We simply replace the stochastic variables by
the values taken in the scenario, and ensure
that the values found for the decision variables
are consistent across scenarios.
Note that certain decision
variables are shared across scenarios. The first stage
decisions are, for example, shared by all
scenarios. The great advantage of this approach
is that we can use conventional constraint solvers
to solve stochastic constraint
programs. We do not need to implement specialized
solvers. Of course, there is a price to pay
as the number of scenarios grows exponentially with
the number of stages. However, our results show
that a scenario-based approach is feasible
for many problems. Indeed, we observe much better performance
using this approach on the production planning example
introduced in \cite{wecai2002}. In addition, as we discuss
later, we have developed a number of techniques like Latin hypercube sampling
to reduce the number of scenarios considered.

\section{Stochastic OPL}

We have implemented this framework on top
of the OPL constraint modelling language
\cite{hent1}.
An OPL model consists of two parts: a set of declarations,
followed by an instruction.
Declarations define the data types, (input) data
and the (decision) variables.
An OPL instruction is either to satisfy
a set of constraints or to maximize/minimize an
objective function subject to a set of constraints.
We have extended the declarations to
include the declaration of stochastic variables,
and the instructions to include chance constraints,
and a range of new goals like maximizing the
expectation of an objective function.

\subsection{Variable declaration}

We now declare both decision and
stochastic variables.
Stochastic variables are set according to a probability
distribution using a command of the form:
\begin{verbatim}
 stoch <Type> <Id> <Dist>;
\end{verbatim}
Where {\tt <Type>} is (as with decision
variables) a data type (e.g. a range of values,
or an enumerated list of values),
{\tt <Id>} is (as with decision
variables) the variable name, and
{\tt <Dist>} defines the probability distribution
of the stochastic variable(s).
Probability distributions include {\tt uniform},
{\tt poisson(lambda)}, and user defined
via a list of (not necessarily normalized)
values. Other types of distribution can be supported
as needed.
We insist that stochastic variables are arrays, with
the last index describing the stage.
Here are some examples:
\begin{verbatim}
 stoch 0..1 market[Years] uniform;
 stoch 100..102 demand[Quarter] {1,2,3};
\end{verbatim}
In the first, we have a 0/1 variable in each year which
takes either value with equal probability.
In the last, we have a demand variable for each quarter, which
takes the value 100 in 1 out of 6 cases,
101 in 2 out 6 cases, and 102 in the remaining
3 cases.

\subsection{Constraint posting}

We can post both hard constraints (as in OPL) and
chance constraints.
Chance constraints hold in some
but not necessarily all scenarios.
They are posted using a command of the form:
\begin{verbatim}
 prob(<Constraint>) <ArithOp> <Expr>;
\end{verbatim}
Where {\tt <Constraint>} is any
OPL constraint,
{\tt <ArithOp>} is any of the
arithmetically comparison operations ({\tt =},{\tt <>},{\tt <},{\tt >},
{\tt <=}, or {\tt >=})
and
{\tt <Expr>} is any arithmetic expression (it
may contain decision variables or may
just be a rational or a float in the range 0 to 1).
For example, the following command specifies the chance constraint
that in each quarter the demand (a stochastic variable)
does not exceed the production (a decision variable) plus
the stock carried forward in each quarter
(this auxiliary is modelled, as in conventional
constraint programming, by a decision variable)
with 80\% probability:
\begin{verbatim}
 forall(i in 1..n)
    prob(demand[i] <=
           production[i]+stock[i])
       >= 0.80;
\end{verbatim}
Constraints which are not chance constraints
are hard and have to hold
in all possible scenarios.
For example, the stock carried forwards
is computed via the
hard constraint:
\begin{verbatim}
 forall(i in 1..n)
    stock[i+1] = stock[i] + production[i]
                          - demand[i];
\end{verbatim}

\subsection{Optimization}

Stochastic OPL supports both stochastic constraint satisfaction and
optimization problems. We can maximize or
minimize the expectation of an objective function. For example,
in the book production example of \cite{wecai2002}, we can
minimize the expected cost of storing surplus books. Each book
costs \$1 per quarter to store. This can be specified by the
following (partial) model:
\begin{verbatim}
 minimize expected(cost)
 subject to
   cost =
     sum(i in 1..n) max(stock[i+1],0);
   forall(i in 1..n)
     stock[i+1] = stock[i] + production[i]
                           - demand[i];
\end{verbatim}
Stochastic OPL also supports a number of other optimization
goals. For example:
\begin{verbatim}
 minimize spread(profit)
 maximize downside(profit)
 minimize upside(cost)
\end{verbatim}
The spread is the difference between the value of the objective
function in the best and worst scenarios, whilst the downside
(upside) is the minimum (maximum) objective function value a
possible scenario may take.

\section{Compilation of stochastic OPL}

These stochastic extensions are compiled down into conventional
(non-stochastic) OPL models automatically by exploiting the
scenario-based semantics. The compiler is written in Lex and Yacc,
with a graphical interface in Visual C++. Compilation involves
replacing stochastic variables by their possible values, and
decision variables by a ragged array of decision variables, one
for each possible scenario. Consider again the chance constraint:
\begin{verbatim}
  prob(
    demand[i] <= production[i]+stock[i])
       >= 0.80;
\end{verbatim}
This is compiled into a sum constraint
of the form:
\begin{verbatim}
 sum(j in Scenarios) p[j]*
   (demand[i,j] <=
      production[i,j]+stock[i,j])
    >= 0.80;
\end{verbatim}
Where \verb+Scenarios+ is the set of scenarios,
\verb+p[j]+ is the probability of scenario \verb+j+,
\verb+demand[i,j]+ is the demand in scenario
\verb+j+ and quarter \verb+i+, etc. Note that the
bracketing of the inequality reifies the constraint so that it
takes the value 1 if satisfied and 0 otherwise.

Hard constraints are also transformed. Consider, for example, the hard constraint:
\begin{verbatim}
 wealth[t] = bonds[t] + stocks[t];
\end{verbatim}
This is compiled into a forall constraint of the form:
\begin{verbatim}
 forall(j in Scenarios)
   wealth[t,j] = bonds[t,j] + stocks[t,j];
\end{verbatim}
Where \verb+wealth[t,j]+ is the wealth at time \verb+t+ in scenario \verb+j+, etc. Maximization and minimization instructions are also transformed. Consider, for example, the optimization instruction:
\begin{verbatim}
 maximize expected(wealth[n])
 subject to ...
\end{verbatim}
This is compiled into an instruction of the form:
\begin{verbatim}
 maximize sum(j in Scenarios)
   p[j]*wealth[n,t]
 subject to ...
\end{verbatim}
The rest of the
stochastic OPL model is transformed in a similar manner.

\section{Value of information and stochastic solutions}

For stochastic
optimization problems, we compute
two statistics which quantify the importance of
randomness.
The value of a stochastic solution (VSS) is the difference
in the objective function for the stochastic problem
(call it the stochastic solution, SS) and the objective value for the deterministic
problem computed by replacing stochastic variables by their
expectations (call it the expected value solution, EVS):
$VSS  =  SS - EVS$.
This computes the benefit of knowing the
distributions of the stochastic variables.
Clearly, VSS is non-negative.
We also compute the expected value of the wait-and-see solution (WSS).
To calculate this, we give the stochastic variables
values according to their probability distributions,
and then find the best values for the decision variables.
The difference between WSS and SS is the expected value
of perfect information (EVPI):
$EVPI  =  WSS - SS$.
This measures how much more you can expect to win if
you have perfect information about the stochastic
components of the problem. In other words, EVPI
measures the value of knowing the future with
certainty. This is therefore the most that should
be spent in gathering information about the
uncertain world.

\section{Scenario reduction}

One problem with a scenario-based approach
is the large number of scenarios, each of which introduces new
decision variables. We have therefore implemented several
techniques to reduce the number of scenarios. The simplest
is to consider just a single scenario in which
stochastic variables take their expected values. This is supported
with the command:
\begin{verbatim}
 scenario expected;
\end{verbatim}
The user may also be content to consider just the most probable
scenarios and ignore rare events. We support this with the
command:
\begin{verbatim}
 scenario top <Num>;
\end{verbatim}
Another option is to use Monte Carlo sampling. The user can
specify the number of scenarios to sample using a command of the
form:
\begin{verbatim}
 scenario sample <Num>;
\end{verbatim}
The probability distributions of the stochastic variables is used
to bias the construction of these scenarios.

We also implemented one of the best sampling methods from
experimental design, and one of the best scenario reduction
methods from operations research. Latin hypercube sampling
\cite{mckay1}, ensures that a range of values for a variable are
sampled. Suppose we want $n$ sample scenarios. We divide the unit
interval into $n$ intervals, and sample a value for each
stochastic variable whose cumulative probability occurs in each of
these interval. We then construct $n$ sample scenarios from these
values, enforcing the condition that the samples use each value
for each stochastic variable exactly once. More precisely, let
$f_i(a)$ be the cumulative probability that $X_i$ takes the value
$a$ or less, $\pi_i(j)$ be the $j$th element of a random
permutation $\pi_i$ of the integers $\{0,\ldots,n-1\}$, and $r$ be
a random number uniformly drawn from $[0,1]$. Then, the $j$th
Latin hypercube sample value for the stochastic variable $X_i$ is:
$$ f_i^{-1}(\frac{ \pi_i(j) + r}{n}) $$
Finally, we implemented a scenario reduction method used in
stochastic programming due to Dupacova, Growe-Kuska and Romisch
\cite{dupacova1}. They report power production planning problems on
which this method offers 90\% accuracy sampling 50\% of the
scenarios and 50\% accuracy sampling just 2\% of the scenarios.
The method heuristically deletes scenarios to approximate as
closely as possible the original scenarios according to a
Fortet-Mourier metric on the stochastic parameter space.

\section{Some examples \label{examples}}

To illustrate the potential of this framework for decision making
under uncertainty, we now
describe a wide range of problems that we
have modelled. In the first problem, we compare
a scenario-based approach to the previous tree search methods
for solving stochastic constraint satisfaction problems.
In the next three problems, we illustrate
the effectiveness of the different scenario
reduction techniques.

\subsection{Production planning \label{bookprod}}

\begin{table*}[htb]
\centering
\begin{tabular}{|c|rr|rr|rrr|}\hline
& \multicolumn{2}{c|}{Backtracking (BT)} & \multicolumn{2}{c|}{Forward
Checking (FC)} & \multicolumn{3}{c|}{Scenario-Based (SB)} \\ No. Stages &
\multicolumn{1}{c}{Nodes} & CPU/sec & \multicolumn{1}{c}{Nodes} &
CPU/sec & Failure & Choice Points & CPU/sec
\\\hline
1 & 28 & 0.01 & 10 & 0.01 & 4 & 5 & 0.00  \\
2 & 650 & 0.09 & 148 & 0.03 & 4 & 8 & 0.02 \\
3 & 17,190 & 2.72 & 3,604 & 0.76 & 8 & 24 & 0.16 \\
4 & 510,356 & 83.81 & 95,570 & 19.07 & 42 & 125 & 1.53 \\
5 & 15,994,856 & 3,245.99 & 2,616,858 & 509.95 & 218 & 690 & 18.52 \\
6 & -- & -- & -- & -- & 1260 & 4035 & 474.47 \\\hline
\end{tabular}
\caption{A Comparison of BT, FC and SB Approaches on the Book
Production Problem (Sec.\ref{bookprod}) \label{table1}}
\end{table*}

This problem comes from \cite{wecai2002}. The results in Table
\ref{table1} show that a scenario-based approach offers much
better performance on this problem than the forward checking or
backtracking tree search algorithms also introduced in this paper. The problem
involves planning production over $m$ quarters. In each quarter,
we expect to sell between 100 and 105 copies of a book. To keep
customers happy, we want to satisfy demand over all $m$ quarters
with 80\% probability. This problem is modelled by an $m$ stage
stochastic CSP. There are $m$ decision variables, $x_i$
representing production in each quarter. There are also $m$
stochastic variables, $y_i$ representing demand in each quarter.
To limit stock carried forward, we use a simple heuristic which
picks the smallest possible values for the decision variables. An
alternative is to convert the problem into an optimization problem
with a cost to keep books in store. We do not explore this option
here, though it is very easy to implement in stochastic OPL, as we
cannot then compare our results with those of the forward checking
or backtracking algorithms from \cite{wecai2002}.

\subsection{Portfolio management}

This portfolio management problem of
\cite{birge1} can be
modelled as a stochastic COP.
Suppose we have $\$P$ to invest in any of
$I$ investments and we wish to
exceed a wealth of $\$G$ after $t$ investment periods. To
calculate the utility, we suppose that
exceeding $\$G$ is equivalent to an
income of $q\%$ of the excess while not meeting the goal is
equivalent to borrowing at a cost $r\%$ of the amount short. This
defines a concave utility function for $r>q$. The
uncertainty in this problem is the rate of return, which is a
random variable, on each investment in each period. The objective
is to determine the optimal investment strategy, which maximizes
the investor's expected utility.

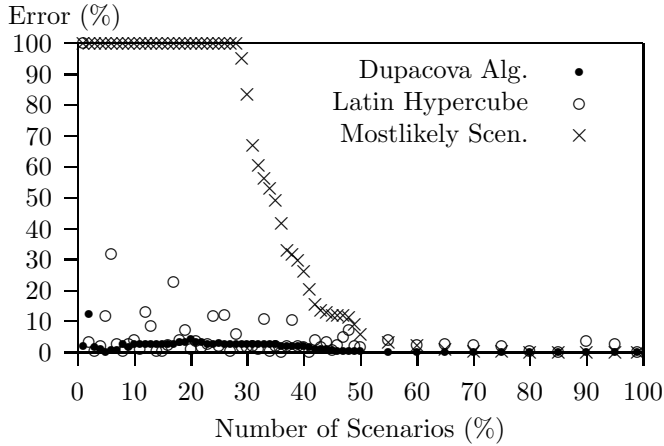
\begin{figure}[htb]
\setlength{\unitlength}{0.240900pt}
\ifx\plotpoint\undefined\newsavebox{\plotpoint}\fi
\sbox{\plotpoint}{\rule[-0.200pt]{0.400pt}{0.400pt}}%
\begin{picture}(1049,629)(0,0)
\font\gnuplot=cmr10 at 10pt
\gnuplot
\sbox{\plotpoint}{\rule[-0.200pt]{0.400pt}{0.400pt}}%
\put(100,143){\makebox(0,0)[r]{0}}
\put(120.0,143.0){\rule[-0.200pt]{4.818pt}{0.400pt}}
\put(100,192){\makebox(0,0)[r]{10}}
\put(120.0,192.0){\rule[-0.200pt]{4.818pt}{0.400pt}}
\put(100,240){\makebox(0,0)[r]{20}}
\put(120.0,240.0){\rule[-0.200pt]{4.818pt}{0.400pt}}
\put(100,289){\makebox(0,0)[r]{30}}
\put(120.0,289.0){\rule[-0.200pt]{4.818pt}{0.400pt}}
\put(100,337){\makebox(0,0)[r]{40}}
\put(120.0,337.0){\rule[-0.200pt]{4.818pt}{0.400pt}}
\put(100,386){\makebox(0,0)[r]{50}}
\put(120.0,386.0){\rule[-0.200pt]{4.818pt}{0.400pt}}
\put(100,435){\makebox(0,0)[r]{60}}
\put(120.0,435.0){\rule[-0.200pt]{4.818pt}{0.400pt}}
\put(100,483){\makebox(0,0)[r]{70}}
\put(120.0,483.0){\rule[-0.200pt]{4.818pt}{0.400pt}}
\put(100,532){\makebox(0,0)[r]{80}}
\put(120.0,532.0){\rule[-0.200pt]{4.818pt}{0.400pt}}
\put(100,580){\makebox(0,0)[r]{90}}
\put(120.0,580.0){\rule[-0.200pt]{4.818pt}{0.400pt}}
\put(100,629){\makebox(0,0)[r]{100}}
\put(120.0,629.0){\rule[-0.200pt]{4.818pt}{0.400pt}}
\put(140,82){\makebox(0,0){0}}
\put(140.0,123.0){\rule[-0.200pt]{0.400pt}{4.818pt}}
\put(229,82){\makebox(0,0){10}}
\put(229.0,123.0){\rule[-0.200pt]{0.400pt}{4.818pt}}
\put(318,82){\makebox(0,0){20}}
\put(318.0,123.0){\rule[-0.200pt]{0.400pt}{4.818pt}}
\put(406,82){\makebox(0,0){30}}
\put(406.0,123.0){\rule[-0.200pt]{0.400pt}{4.818pt}}
\put(495,82){\makebox(0,0){40}}
\put(495.0,123.0){\rule[-0.200pt]{0.400pt}{4.818pt}}
\put(584,82){\makebox(0,0){50}}
\put(584.0,123.0){\rule[-0.200pt]{0.400pt}{4.818pt}}
\put(673,82){\makebox(0,0){60}}
\put(673.0,123.0){\rule[-0.200pt]{0.400pt}{4.818pt}}
\put(762,82){\makebox(0,0){70}}
\put(762.0,123.0){\rule[-0.200pt]{0.400pt}{4.818pt}}
\put(850,82){\makebox(0,0){80}}
\put(850.0,123.0){\rule[-0.200pt]{0.400pt}{4.818pt}}
\put(939,82){\makebox(0,0){90}}
\put(939.0,123.0){\rule[-0.200pt]{0.400pt}{4.818pt}}
\put(1028,82){\makebox(0,0){100}}
\put(1028.0,123.0){\rule[-0.200pt]{0.400pt}{4.818pt}}
\put(140.0,143.0){\rule[-0.200pt]{213.919pt}{0.400pt}}
\put(1028.0,143.0){\rule[-0.200pt]{0.400pt}{117.077pt}}
\put(140.0,629.0){\rule[-0.200pt]{213.919pt}{0.400pt}}
\put(584,21){\makebox(0,0){Number of Scenarios (\%)}}
\put(120,670){\makebox(0,0){Error (\%)}}
\put(140.0,143.0){\rule[-0.200pt]{0.400pt}{117.077pt}}
\sbox{\plotpoint}{\rule[-0.500pt]{1.000pt}{1.000pt}}%
\put(848,584){\makebox(0,0)[r]{Dupacova Alg.}}
\put(1019,143){\circle*{12}}
\put(984,143){\circle*{12}}
\put(939,143){\circle*{12}}
\put(895,143){\circle*{12}}
\put(850,143){\circle*{12}}
\put(806,143){\circle*{12}}
\put(762,143){\circle*{12}}
\put(717,143){\circle*{12}}
\put(673,144){\circle*{12}}
\put(628,144){\circle*{12}}
\put(584,145){\circle*{12}}
\put(575,145){\circle*{12}}
\put(566,146){\circle*{12}}
\put(557,145){\circle*{12}}
\put(548,146){\circle*{12}}
\put(540,147){\circle*{12}}
\put(531,148){\circle*{12}}
\put(522,148){\circle*{12}}
\put(513,148){\circle*{12}}
\put(504,152){\circle*{12}}
\put(495,153){\circle*{12}}
\put(486,153){\circle*{12}}
\put(477,153){\circle*{12}}
\put(469,153){\circle*{12}}
\put(460,153){\circle*{12}}
\put(451,156){\circle*{12}}
\put(442,156){\circle*{12}}
\put(433,156){\circle*{12}}
\put(424,157){\circle*{12}}
\put(415,157){\circle*{12}}
\put(406,157){\circle*{12}}
\put(398,157){\circle*{12}}
\put(389,157){\circle*{12}}
\put(380,156){\circle*{12}}
\put(371,156){\circle*{12}}
\put(362,158){\circle*{12}}
\put(353,157){\circle*{12}}
\put(344,157){\circle*{12}}
\put(335,159){\circle*{12}}
\put(326,159){\circle*{12}}
\put(318,164){\circle*{12}}
\put(309,160){\circle*{12}}
\put(300,159){\circle*{12}}
\put(291,157){\circle*{12}}
\put(282,157){\circle*{12}}
\put(273,156){\circle*{12}}
\put(264,156){\circle*{12}}
\put(255,156){\circle*{12}}
\put(247,156){\circle*{12}}
\put(238,156){\circle*{12}}
\put(229,156){\circle*{12}}
\put(220,151){\circle*{12}}
\put(211,156){\circle*{12}}
\put(202,147){\circle*{12}}
\put(193,147){\circle*{12}}
\put(184,143){\circle*{12}}
\put(176,148){\circle*{12}}
\put(167,151){\circle*{12}}
\put(158,203){\circle*{12}}
\put(149,153){\circle*{12}}
\put(928,584){\circle*{12}}
\sbox{\plotpoint}{\rule[-0.200pt]{0.400pt}{0.400pt}}%
\put(848,534){\makebox(0,0)[r]{Latin Hypercube}}
\put(1019,143){\circle{18}}
\put(984,156){\circle{18}}
\put(939,161){\circle{18}}
\put(895,144){\circle{18}}
\put(850,145){\circle{18}}
\put(806,153){\circle{18}}
\put(762,154){\circle{18}}
\put(717,156){\circle{18}}
\put(673,154){\circle{18}}
\put(628,162){\circle{18}}
\put(584,152){\circle{18}}
\put(575,153){\circle{18}}
\put(566,178){\circle{18}}
\put(557,168){\circle{18}}
\put(548,155){\circle{18}}
\put(540,147){\circle{18}}
\put(531,159){\circle{18}}
\put(522,151){\circle{18}}
\put(513,162){\circle{18}}
\put(504,143){\circle{18}}
\put(495,152){\circle{18}}
\put(486,153){\circle{18}}
\put(477,194){\circle{18}}
\put(469,153){\circle{18}}
\put(460,143){\circle{18}}
\put(451,153){\circle{18}}
\put(442,145){\circle{18}}
\put(433,195){\circle{18}}
\put(424,148){\circle{18}}
\put(415,153){\circle{18}}
\put(406,151){\circle{18}}
\put(398,152){\circle{18}}
\put(389,172){\circle{18}}
\put(380,145){\circle{18}}
\put(371,202){\circle{18}}
\put(362,153){\circle{18}}
\put(353,201){\circle{18}}
\put(344,156){\circle{18}}
\put(335,152){\circle{18}}
\put(326,161){\circle{18}}
\put(318,149){\circle{18}}
\put(309,178){\circle{18}}
\put(300,162){\circle{18}}
\put(291,254){\circle{18}}
\put(282,154){\circle{18}}
\put(273,146){\circle{18}}
\put(264,146){\circle{18}}
\put(255,185){\circle{18}}
\put(247,207){\circle{18}}
\put(238,149){\circle{18}}
\put(229,162){\circle{18}}
\put(220,157){\circle{18}}
\put(211,145){\circle{18}}
\put(202,157){\circle{18}}
\put(193,298){\circle{18}}
\put(184,200){\circle{18}}
\put(176,153){\circle{18}}
\put(167,146){\circle{18}}
\put(158,159){\circle{18}}
\put(149,629){\circle{18}}
\put(928,534){\circle{18}}
\sbox{\plotpoint}{\rule[-0.500pt]{1.000pt}{1.000pt}}%
\put(848,484){\makebox(0,0)[r]{Mostlikely Scen.}}
\put(1019,143){\makebox(0,0){$\times$}}
\put(984,143){\makebox(0,0){$\times$}}
\put(939,143){\makebox(0,0){$\times$}}
\put(895,143){\makebox(0,0){$\times$}}
\put(850,143){\makebox(0,0){$\times$}}
\put(806,144){\makebox(0,0){$\times$}}
\put(762,145){\makebox(0,0){$\times$}}
\put(717,148){\makebox(0,0){$\times$}}
\put(673,154){\makebox(0,0){$\times$}}
\put(628,159){\makebox(0,0){$\times$}}
\put(584,172){\makebox(0,0){$\times$}}
\put(575,188){\makebox(0,0){$\times$}}
\put(566,198){\makebox(0,0){$\times$}}
\put(557,202){\makebox(0,0){$\times$}}
\put(548,202){\makebox(0,0){$\times$}}
\put(540,202){\makebox(0,0){$\times$}}
\put(531,206){\makebox(0,0){$\times$}}
\put(522,209){\makebox(0,0){$\times$}}
\put(513,219){\makebox(0,0){$\times$}}
\put(504,242){\makebox(0,0){$\times$}}
\put(495,270){\makebox(0,0){$\times$}}
\put(486,288){\makebox(0,0){$\times$}}
\put(477,297){\makebox(0,0){$\times$}}
\put(469,304){\makebox(0,0){$\times$}}
\put(460,346){\makebox(0,0){$\times$}}
\put(451,382){\makebox(0,0){$\times$}}
\put(442,401){\makebox(0,0){$\times$}}
\put(433,417){\makebox(0,0){$\times$}}
\put(424,437){\makebox(0,0){$\times$}}
\put(415,468){\makebox(0,0){$\times$}}
\put(406,548){\makebox(0,0){$\times$}}
\put(398,606){\makebox(0,0){$\times$}}
\put(389,629){\makebox(0,0){$\times$}}
\put(380,629){\makebox(0,0){$\times$}}
\put(371,629){\makebox(0,0){$\times$}}
\put(362,629){\makebox(0,0){$\times$}}
\put(353,629){\makebox(0,0){$\times$}}
\put(344,629){\makebox(0,0){$\times$}}
\put(335,629){\makebox(0,0){$\times$}}
\put(326,629){\makebox(0,0){$\times$}}
\put(318,629){\makebox(0,0){$\times$}}
\put(309,629){\makebox(0,0){$\times$}}
\put(300,629){\makebox(0,0){$\times$}}
\put(291,629){\makebox(0,0){$\times$}}
\put(282,629){\makebox(0,0){$\times$}}
\put(273,629){\makebox(0,0){$\times$}}
\put(264,629){\makebox(0,0){$\times$}}
\put(255,629){\makebox(0,0){$\times$}}
\put(247,629){\makebox(0,0){$\times$}}
\put(238,629){\makebox(0,0){$\times$}}
\put(229,629){\makebox(0,0){$\times$}}
\put(220,629){\makebox(0,0){$\times$}}
\put(211,629){\makebox(0,0){$\times$}}
\put(202,629){\makebox(0,0){$\times$}}
\put(193,629){\makebox(0,0){$\times$}}
\put(184,629){\makebox(0,0){$\times$}}
\put(176,629){\makebox(0,0){$\times$}}
\put(167,629){\makebox(0,0){$\times$}}
\put(158,629){\makebox(0,0){$\times$}}
\put(149,629){\makebox(0,0){$\times$}}
\put(928,484){\makebox(0,0){$\times$}}
\end{picture}

\caption{Portfolio Management}
\label{fig1}
\end{figure}

The problem has 8 stages and 5760 scenarios.
To compare the effectiveness of the different scenario reduction algorithms,
we adopt a two step procedure. In the first step, the scenario reduced
problem is solved and the first period's decision is observed.
We then solve the full-size (non scenario reduced) problem to optimality
with this first decision fixed. The difference
between the objective values of these two solutions is
normalized by the range [optimal solution, observed worst solution]
to give a normalized error for committing to the scenario reduced first
decision. In Fig. \ref{fig1}, we see that Dupacova et al's algorithm
is very effective, that Latin hypercube sampling is a small
distance behind, and both are far ahead of the most likely
scenario method (which requires approximately half the scenarios
before the first decision is made correctly).

\subsection{Yield management}

\begin{figure}
\setlength{\unitlength}{0.240900pt}
\ifx\plotpoint\undefined\newsavebox{\plotpoint}\fi
\sbox{\plotpoint}{\rule[-0.200pt]{0.400pt}{0.400pt}}%
\begin{picture}(1049,629)(0,0)
\font\gnuplot=cmr10 at 10pt
\gnuplot
\sbox{\plotpoint}{\rule[-0.200pt]{0.400pt}{0.400pt}}%
\put(100,143){\makebox(0,0)[r]{0}}
\put(120.0,143.0){\rule[-0.200pt]{4.818pt}{0.400pt}}
\put(100,192){\makebox(0,0)[r]{10}}
\put(120.0,192.0){\rule[-0.200pt]{4.818pt}{0.400pt}}
\put(100,240){\makebox(0,0)[r]{20}}
\put(120.0,240.0){\rule[-0.200pt]{4.818pt}{0.400pt}}
\put(100,289){\makebox(0,0)[r]{30}}
\put(120.0,289.0){\rule[-0.200pt]{4.818pt}{0.400pt}}
\put(100,337){\makebox(0,0)[r]{40}}
\put(120.0,337.0){\rule[-0.200pt]{4.818pt}{0.400pt}}
\put(100,386){\makebox(0,0)[r]{50}}
\put(120.0,386.0){\rule[-0.200pt]{4.818pt}{0.400pt}}
\put(100,435){\makebox(0,0)[r]{60}}
\put(120.0,435.0){\rule[-0.200pt]{4.818pt}{0.400pt}}
\put(100,483){\makebox(0,0)[r]{70}}
\put(120.0,483.0){\rule[-0.200pt]{4.818pt}{0.400pt}}
\put(100,532){\makebox(0,0)[r]{80}}
\put(120.0,532.0){\rule[-0.200pt]{4.818pt}{0.400pt}}
\put(100,580){\makebox(0,0)[r]{90}}
\put(120.0,580.0){\rule[-0.200pt]{4.818pt}{0.400pt}}
\put(100,629){\makebox(0,0)[r]{100}}
\put(120.0,629.0){\rule[-0.200pt]{4.818pt}{0.400pt}}
\put(140,82){\makebox(0,0){0}}
\put(140.0,123.0){\rule[-0.200pt]{0.400pt}{4.818pt}}
\put(229,82){\makebox(0,0){10}}
\put(229.0,123.0){\rule[-0.200pt]{0.400pt}{4.818pt}}
\put(318,82){\makebox(0,0){20}}
\put(318.0,123.0){\rule[-0.200pt]{0.400pt}{4.818pt}}
\put(406,82){\makebox(0,0){30}}
\put(406.0,123.0){\rule[-0.200pt]{0.400pt}{4.818pt}}
\put(495,82){\makebox(0,0){40}}
\put(495.0,123.0){\rule[-0.200pt]{0.400pt}{4.818pt}}
\put(584,82){\makebox(0,0){50}}
\put(584.0,123.0){\rule[-0.200pt]{0.400pt}{4.818pt}}
\put(673,82){\makebox(0,0){60}}
\put(673.0,123.0){\rule[-0.200pt]{0.400pt}{4.818pt}}
\put(762,82){\makebox(0,0){70}}
\put(762.0,123.0){\rule[-0.200pt]{0.400pt}{4.818pt}}
\put(850,82){\makebox(0,0){80}}
\put(850.0,123.0){\rule[-0.200pt]{0.400pt}{4.818pt}}
\put(939,82){\makebox(0,0){90}}
\put(939.0,123.0){\rule[-0.200pt]{0.400pt}{4.818pt}}
\put(1028,82){\makebox(0,0){100}}
\put(1028.0,123.0){\rule[-0.200pt]{0.400pt}{4.818pt}}
\put(140.0,143.0){\rule[-0.200pt]{213.919pt}{0.400pt}}
\put(1028.0,143.0){\rule[-0.200pt]{0.400pt}{117.077pt}}
\put(140.0,629.0){\rule[-0.200pt]{213.919pt}{0.400pt}}
\put(584,21){\makebox(0,0){Number of Scenarios (\%)}}
\put(120,670){\makebox(0,0){Error (\%)}}
\put(140.0,143.0){\rule[-0.200pt]{0.400pt}{117.077pt}}
\sbox{\plotpoint}{\rule[-0.500pt]{1.000pt}{1.000pt}}%
\put(848,584){\makebox(0,0)[r]{Dupacova Alg.}}
\put(1019,143){\circle*{12}}
\put(984,143){\circle*{12}}
\put(939,143){\circle*{12}}
\put(895,143){\circle*{12}}
\put(850,143){\circle*{12}}
\put(806,143){\circle*{12}}
\put(762,143){\circle*{12}}
\put(717,143){\circle*{12}}
\put(673,143){\circle*{12}}
\put(628,143){\circle*{12}}
\put(584,143){\circle*{12}}
\put(540,143){\circle*{12}}
\put(495,143){\circle*{12}}
\put(451,143){\circle*{12}}
\put(406,143){\circle*{12}}
\put(362,143){\circle*{12}}
\put(318,143){\circle*{12}}
\put(273,143){\circle*{12}}
\put(229,143){\circle*{12}}
\put(184,143){\circle*{12}}
\put(149,149){\circle*{12}}
\put(928,584){\circle*{12}}
\sbox{\plotpoint}{\rule[-0.200pt]{0.400pt}{0.400pt}}%
\put(848,534){\makebox(0,0)[r]{Latin Hypercube}}
\put(1019,143){\circle{18}}
\put(984,143){\circle{18}}
\put(939,143){\circle{18}}
\put(895,143){\circle{18}}
\put(850,143){\circle{18}}
\put(806,143){\circle{18}}
\put(762,143){\circle{18}}
\put(717,143){\circle{18}}
\put(673,143){\circle{18}}
\put(628,143){\circle{18}}
\put(584,143){\circle{18}}
\put(540,143){\circle{18}}
\put(495,143){\circle{18}}
\put(451,143){\circle{18}}
\put(406,143){\circle{18}}
\put(362,143){\circle{18}}
\put(318,143){\circle{18}}
\put(273,143){\circle{18}}
\put(229,143){\circle{18}}
\put(184,143){\circle{18}}
\put(149,143){\circle{18}}
\put(1019,143){\circle{18}}
\put(928,534){\circle{18}}
\sbox{\plotpoint}{\rule[-0.500pt]{1.000pt}{1.000pt}}%
\put(848,484){\makebox(0,0)[r]{Mostlikely Scen.}}
\put(1019,143){\makebox(0,0){$\times$}}
\put(984,143){\makebox(0,0){$\times$}}
\put(939,143){\makebox(0,0){$\times$}}
\put(895,143){\makebox(0,0){$\times$}}
\put(850,143){\makebox(0,0){$\times$}}
\put(806,143){\makebox(0,0){$\times$}}
\put(762,143){\makebox(0,0){$\times$}}
\put(717,143){\makebox(0,0){$\times$}}
\put(673,143){\makebox(0,0){$\times$}}
\put(628,143){\makebox(0,0){$\times$}}
\put(584,143){\makebox(0,0){$\times$}}
\put(540,143){\makebox(0,0){$\times$}}
\put(495,151){\makebox(0,0){$\times$}}
\put(451,154){\makebox(0,0){$\times$}}
\put(406,241){\makebox(0,0){$\times$}}
\put(362,243){\makebox(0,0){$\times$}}
\put(318,243){\makebox(0,0){$\times$}}
\put(273,243){\makebox(0,0){$\times$}}
\put(229,243){\makebox(0,0){$\times$}}
\put(184,592){\makebox(0,0){$\times$}}
\put(149,629){\makebox(0,0){$\times$}}
\put(928,484){\makebox(0,0){$\times$}}
\end{picture}

\caption{Agricultural Yield Management}
\label{fig2}
\end{figure}
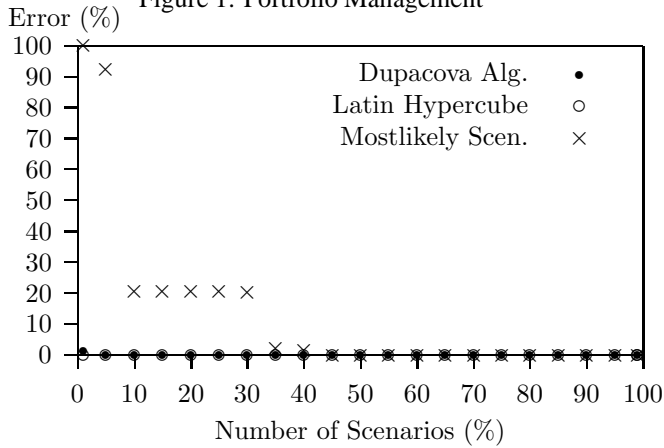

Farmers must deal with uncertainty since weather and many other
factors affect crop yields. In this
example (also taken from \cite{birge1}), we must
decide on how many acres of his fields to devote to various crops
before the planting season. A certain amount of each crop is
required for cattle feed, which can be purchased from a wholesaler
if not raised on the farm. Any crop in excess of cattle feed can
be sold up to the EU quota; any amount in excess of this
quota will be sold at a low price. Crop yields are
uncertain, depending upon weather conditions during the growing
season. This problem has 4 stages and 10,000 scenarios.
In Fig. \ref{fig2}, we again see that Dupacova et al's algorithm
and Latin hypercube sampling are very effective,
and both are far ahead of the most likely
scenario method (which requires approximately one third the scenarios
before the first decision is made correctly).

\subsection{Production/Inventory control \label{proinv}}

\begin{figure}
\setlength{\unitlength}{0.240900pt}
\ifx\plotpoint\undefined\newsavebox{\plotpoint}\fi
\sbox{\plotpoint}{\rule[-0.200pt]{0.400pt}{0.400pt}}%
\begin{picture}(1049,629)(0,0)
\font\gnuplot=cmr10 at 10pt
\gnuplot
\sbox{\plotpoint}{\rule[-0.200pt]{0.400pt}{0.400pt}}%
\put(100,143){\makebox(0,0)[r]{0}}
\put(120.0,143.0){\rule[-0.200pt]{4.818pt}{0.400pt}}
\put(100,192){\makebox(0,0)[r]{10}}
\put(120.0,192.0){\rule[-0.200pt]{4.818pt}{0.400pt}}
\put(100,240){\makebox(0,0)[r]{20}}
\put(120.0,240.0){\rule[-0.200pt]{4.818pt}{0.400pt}}
\put(100,289){\makebox(0,0)[r]{30}}
\put(120.0,289.0){\rule[-0.200pt]{4.818pt}{0.400pt}}
\put(100,337){\makebox(0,0)[r]{40}}
\put(120.0,337.0){\rule[-0.200pt]{4.818pt}{0.400pt}}
\put(100,386){\makebox(0,0)[r]{50}}
\put(120.0,386.0){\rule[-0.200pt]{4.818pt}{0.400pt}}
\put(100,435){\makebox(0,0)[r]{60}}
\put(120.0,435.0){\rule[-0.200pt]{4.818pt}{0.400pt}}
\put(100,483){\makebox(0,0)[r]{70}}
\put(120.0,483.0){\rule[-0.200pt]{4.818pt}{0.400pt}}
\put(100,532){\makebox(0,0)[r]{80}}
\put(120.0,532.0){\rule[-0.200pt]{4.818pt}{0.400pt}}
\put(100,580){\makebox(0,0)[r]{90}}
\put(120.0,580.0){\rule[-0.200pt]{4.818pt}{0.400pt}}
\put(100,629){\makebox(0,0)[r]{100}}
\put(120.0,629.0){\rule[-0.200pt]{4.818pt}{0.400pt}}
\put(140,82){\makebox(0,0){0}}
\put(140.0,123.0){\rule[-0.200pt]{0.400pt}{4.818pt}}
\put(229,82){\makebox(0,0){10}}
\put(229.0,123.0){\rule[-0.200pt]{0.400pt}{4.818pt}}
\put(318,82){\makebox(0,0){20}}
\put(318.0,123.0){\rule[-0.200pt]{0.400pt}{4.818pt}}
\put(406,82){\makebox(0,0){30}}
\put(406.0,123.0){\rule[-0.200pt]{0.400pt}{4.818pt}}
\put(495,82){\makebox(0,0){40}}
\put(495.0,123.0){\rule[-0.200pt]{0.400pt}{4.818pt}}
\put(584,82){\makebox(0,0){50}}
\put(584.0,123.0){\rule[-0.200pt]{0.400pt}{4.818pt}}
\put(673,82){\makebox(0,0){60}}
\put(673.0,123.0){\rule[-0.200pt]{0.400pt}{4.818pt}}
\put(762,82){\makebox(0,0){70}}
\put(762.0,123.0){\rule[-0.200pt]{0.400pt}{4.818pt}}
\put(850,82){\makebox(0,0){80}}
\put(850.0,123.0){\rule[-0.200pt]{0.400pt}{4.818pt}}
\put(939,82){\makebox(0,0){90}}
\put(939.0,123.0){\rule[-0.200pt]{0.400pt}{4.818pt}}
\put(1028,82){\makebox(0,0){100}}
\put(1028.0,123.0){\rule[-0.200pt]{0.400pt}{4.818pt}}
\put(140.0,143.0){\rule[-0.200pt]{213.919pt}{0.400pt}}
\put(1028.0,143.0){\rule[-0.200pt]{0.400pt}{117.077pt}}
\put(140.0,629.0){\rule[-0.200pt]{213.919pt}{0.400pt}}
\put(584,21){\makebox(0,0){Number of Scenarios (\%)}}
\put(120,670){\makebox(0,0){Error (\%)}}
\put(140.0,143.0){\rule[-0.200pt]{0.400pt}{117.077pt}}
\sbox{\plotpoint}{\rule[-0.500pt]{1.000pt}{1.000pt}}%
\put(848,584){\makebox(0,0)[r]{Dupacova Alg.}}
\put(1019,143){\circle*{12}}
\put(984,143){\circle*{12}}
\put(939,143){\circle*{12}}
\put(895,143){\circle*{12}}
\put(850,143){\circle*{12}}
\put(806,143){\circle*{12}}
\put(762,143){\circle*{12}}
\put(717,143){\circle*{12}}
\put(673,143){\circle*{12}}
\put(628,143){\circle*{12}}
\put(584,143){\circle*{12}}
\put(540,143){\circle*{12}}
\put(495,143){\circle*{12}}
\put(451,143){\circle*{12}}
\put(406,143){\circle*{12}}
\put(362,143){\circle*{12}}
\put(318,143){\circle*{12}}
\put(273,143){\circle*{12}}
\put(229,143){\circle*{12}}
\put(184,143){\circle*{12}}
\put(149,143){\circle*{12}}
\put(928,584){\circle*{12}}
\sbox{\plotpoint}{\rule[-0.200pt]{0.400pt}{0.400pt}}%
\put(848,534){\makebox(0,0)[r]{Latin Hypercube}}
\put(1019,143){\circle{18}}
\put(984,143){\circle{18}}
\put(939,143){\circle{18}}
\put(895,143){\circle{18}}
\put(850,143){\circle{18}}
\put(806,143){\circle{18}}
\put(762,143){\circle{18}}
\put(717,143){\circle{18}}
\put(673,143){\circle{18}}
\put(628,143){\circle{18}}
\put(584,143){\circle{18}}
\put(540,143){\circle{18}}
\put(495,187){\circle{18}}
\put(451,143){\circle{18}}
\put(406,143){\circle{18}}
\put(362,187){\circle{18}}
\put(318,143){\circle{18}}
\put(273,143){\circle{18}}
\put(229,143){\circle{18}}
\put(184,143){\circle{18}}
\put(149,166){\circle{18}}
\put(928,534){\circle{18}}
\sbox{\plotpoint}{\rule[-0.500pt]{1.000pt}{1.000pt}}%
\put(848,484){\makebox(0,0)[r]{Mostlikely Scen.}}
\put(1019,143){\makebox(0,0){$\times$}}
\put(984,143){\makebox(0,0){$\times$}}
\put(939,143){\makebox(0,0){$\times$}}
\put(895,143){\makebox(0,0){$\times$}}
\put(850,143){\makebox(0,0){$\times$}}
\put(806,143){\makebox(0,0){$\times$}}
\put(762,143){\makebox(0,0){$\times$}}
\put(717,143){\makebox(0,0){$\times$}}
\put(673,143){\makebox(0,0){$\times$}}
\put(628,143){\makebox(0,0){$\times$}}
\put(584,143){\makebox(0,0){$\times$}}
\put(540,143){\makebox(0,0){$\times$}}
\put(495,143){\makebox(0,0){$\times$}}
\put(451,143){\makebox(0,0){$\times$}}
\put(406,187){\makebox(0,0){$\times$}}
\put(362,187){\makebox(0,0){$\times$}}
\put(318,187){\makebox(0,0){$\times$}}
\put(273,187){\makebox(0,0){$\times$}}
\put(229,187){\makebox(0,0){$\times$}}
\put(184,187){\makebox(0,0){$\times$}}
\put(149,187){\makebox(0,0){$\times$}}
\put(928,484){\makebox(0,0){$\times$}}
\end{picture}

\caption{Production/Inventory Control}
\label{fig3}
\end{figure}
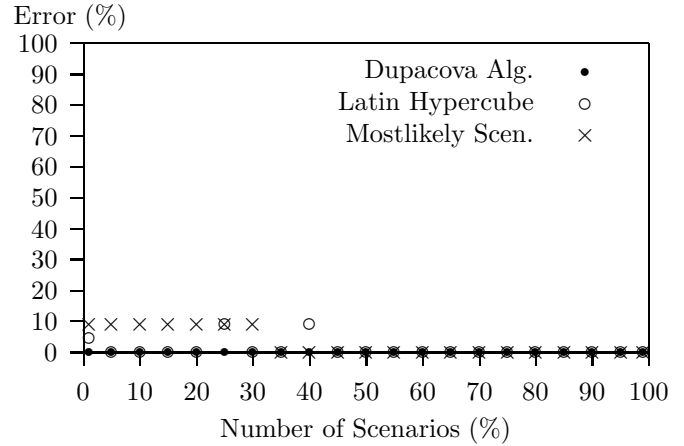

Uncertainty plays a major role in production
and inventory planning. In this simplified
production/inventory planning example, there
is a single product, a single stocking point,
production capacity constraints and stochastic
demand. The objective is to find the minimum expected cost policy.
The cost components take into account holding costs,
backlogging costs, fixed replenishment (or setup) costs and unit
production costs. The optimal policy gives the timing of the
replenishments as well as the order-up-to-levels. Hence, the exact
order quantity can be known only after the realization of the
demand, using the scenario dependent order-up-to-level decisions.
This problem has 5 stages and 1,024 scenarios.
In Fig. \ref{fig3}, we again see that Dupacova et al's algorithm
and Latin hypercube sampling are very effective,
but both are now only a small distance ahead of the most likely
scenario method.

\section{Robust solutions}

Inspired by robust optimization methods in operations research
\cite{robust}, stochastic OPL also allows us
to find robust solutions to stochastic
constraint programs. That is, solutions in which similar decisions
are made in the different scenarios. It will often be impossible
or undesirable for all decision variables to be robust.
We therefore identify those
decision variables whose values we wish to be identical across
scenarios using commands of the form:
\begin{verbatim}
  robust <Var>;
\end{verbatim}

For example, in production/inventory problem of Sec.\ref{proinv}
the decision variables ``order-up-to-levels'' and ``replenishment
periods'' can be declared as robust variables. The values of
these two sets of decision variables are then fixed at the beginning of
the planning horizon. A robust solution
dampens the nervousness of the solution, an area of
very active research in production/inventory management.
As the expected cost of the robust solution is always higher, the
tradeoff between nervousness and cost may have to be taken into account.

\section{Related and future work}

Stochastic constraint programs are closely
related to Markov decision problems (MDPs) \cite{puterman1}.
Stochastic constraint programs can, however, model
problems which lack the Markov property that
the next state and reward depend only on the previous
state and action taken.
The current decision in a stochastic constraint
program will often depend on all
earlier decisions. To model this as an MDP, we would
need an exponential number of states.
Another significant difference is that stochastic constraint
programs by using a scenario-based
interpretation can immediately call upon complex
and powerful constraint propagation techniques.

Stochastic constraint programming was inspired by both stochastic
integer programming and stochastic satisfiability \cite{littman1}.
It is designed to take advantage of some of the best features of
each framework. For example, we are able to write expressive
models using non-linear and global constraints, and to exploit
efficient constraint propagation algorithms. In operations
research, scenarios are used in stochastic programming. Indeed,
the scenario reduction techniques of Dupacova, Growe-Kuska and
Romisch \cite{dupacova1} implemented here are borrowed directly
from stochastic programming.

There are a number of extensions of conventional constraint
satisfaction problem to model constraints that
are uncertain, probabilistic or not necessarily satisfied.
For example, in probabilistic constraint satisfaction
each constraint has a certain probability independent of
all other probabilities of being part of the problem \cite{fargier1}
whilst in semi-ring constraint satisfaction
each tuple in a constraint has a value associated with it \cite{schiex5}.
However, none of these extensions deal with variables
that may have uncertain or probabilistic values.
Stochastic constraint
programming could, however, easily be combined with most
of these techniques.

\section{Conclusions}

To model combinatorial decision problems involving uncertainty and
probability, we have extended the stochastic constraint
programming framework proposed in \cite{wecai2002} along a number
of important dimensions. In particular, we have relaxed the
assumption that stochastic variables are independent, and added
multiple chance constraints as well as a range of objective
functions like maximizing the downside. We have also provided a
new (but equivalent) semantics for stochastic constraint programs
based on scenarios. Based on this semantics, we can compile
stochastic constraint programs down into conventional (non-stochastic)
constraint programs. The advantage of this compilation is that we
can use the full power of existing constraint solvers without any
modification. We have also proposed a number of techniques to
reduce the number of scenarios, and to generate robust solutions.

We have implemented this framework for decision making under
uncertainty in a language called stochastic OPL. This is an
extension of the OPL constraint modelling language \cite{hent1}.
To illustrate the potential of this framework, we have modelled a
wide range of problems in areas as diverse as finance, agriculture
and production. There are many directions for future work. For
example, we want to allow the user to define a limited set of
scenarios that are representative of the whole. As a second
example, we want to explore more sophisticated notions of solution
robustness (e.g. limiting the range of values used by a decision
variable).

\vspace{-2mm}
\section*{Acknowledgements}
This project was funded by EPSRC under GR/R30792, and the Science Foundation
Ireland. We thank the members of the APES Research Group and 4C Lab for their feedback.




\bibliographystyle{named}


\begin{thebibliography}{}

\bibitem[\protect\citeauthoryear{Birge and Louveaux}{1997}]{birge1}
J.~R. Birge and F.~Louveaux.
\newblock {\em Introduction to Stochastic Programming}.
\newblock Springer-Verlag, New York, 1997.

\bibitem[\protect\citeauthoryear{Bistarelli \bgroup \em et al.\egroup
  }{1996}]{schiex5}
S.~Bistarelli, H.~Fargier, U.~Montanari, F.~Rossi, T.~Schiex, and
  G.~Verfaillie.
\newblock Semi-ring based {CSPs} and valued {CSPs}: Basic properties and
  comparison.
\newblock In M.~Jample, E.~Freuder, and M.~Maher, editors, {\em
  Over-Constrained Systems}, pages 111--150. Springer-Verlag, 1996.
\newblock LNCS 1106.

\bibitem[\protect\citeauthoryear{Dupacova \bgroup \em et al.\egroup
  }{2002}]{dupacova1}
J.~Dupacova, N.~Growe-Kuska, and W.~Romisch.
\newblock Scenario reduction in stochastic programming: an approach using
  probability metrics.
\newblock {\em Mathematical Programming}, To appear, 2002.

\bibitem[\protect\citeauthoryear{Fargier and Lang}{1993}]{fargier1}
H.~Fargier and J.~Lang.
\newblock Uncertainty in constraint satisfaction problems: a probabilistic
  approac h.
\newblock In {\em Proceedings of {ECSQARU}}. Springer-Verlag, 1993.
\newblock LNCS 747.

\bibitem[\protect\citeauthoryear{Hentenryck \bgroup \em et al.\egroup
  }{1999}]{hent1}
P.~Van Hentenryck, L.~Michel, L.~Perron, and J-C. Regin.
\newblock Constraint programming in {OPL}.
\newblock In G.~Nadathur, editor, {\em Principles and Practice of Declarative
  Programming}, pages 97--116. Springer-Verlag, 1999.
\newblock Lecture Notes in Computer Science 1702.

\bibitem[\protect\citeauthoryear{Kouvelis and Yu}{1996}]{robust}
P.~Kouvelis and G.~Yu.
\newblock {\em Robust Discrete Optimization and Its Applications}.
\newblock Nonconvex optimization and its applications: volume 14. Kluwer, 1996.

\bibitem[\protect\citeauthoryear{Littman \bgroup \em et al.\egroup
  }{2000}]{littman1}
M.L. Littman, S.M. Majercik, and T.~Pitassi.
\newblock Stochastic {Boolean} satisfiability.
\newblock {\em Journal of Automated Reasoning}, 2000.

\bibitem[\protect\citeauthoryear{McKay \bgroup \em et al.\egroup
  }{1979}]{mckay1}
M.D. McKay, R.J. Beckman, and W.J. Conover.
\newblock {A comparison of three methods for selecting values of input
  variables in the analysis of output from a computer code}.
\newblock {\em {Technometrics}}, 21(2):239--245, 1979.

\bibitem[\protect\citeauthoryear{Puterman}{1994}]{puterman1}
M.L. Puterman.
\newblock {\em Markov decision processes: discrete stochastic dynamic
  programming}.
\newblock John Wiley and Sons, 1994.

\bibitem[\protect\citeauthoryear{Walsh}{2002}]{wecai2002}
Toby Walsh.
\newblock Stochastic constraint programming.
\newblock In {\em Proceedings of the 15th ECAI}. European Conference on
  Artificial Intelligence, IOS Press, 2002.

\end{thebibliography}

\end{document}